\documentclass[letterpaper, 10 pt, conference]{ieeeconf}  

\IEEEoverridecommandlockouts                              

\usepackage{graphicx} 
\usepackage{amsmath} 
\usepackage{amssymb}  
\usepackage{bm}
\usepackage{capt-of}
\usepackage[ruled]{algorithm2e}
\usepackage{algpseudocode}
\usepackage{booktabs}
\usepackage{textcomp}
\usepackage{lipsum}
\usepackage[table]{xcolor}
\usepackage{array,colortbl}
\usepackage[noadjust]{cite}

\usepackage{xcolor}

\usepackage{array}
\newcolumntype{P}[1]{>{\centering\arraybackslash}p{#1}}

\usepackage{hyperref}
\hypersetup{
    colorlinks=true,
    linkcolor=black,
    filecolor=magenta,      
    urlcolor=cyan,
    citecolor=black,
}

\title{\LARGE \bf
\vspace{10pt}
Grasp EveryThing (GET): 1-DoF, 3-Fingered Gripper \\ with Tactile Sensing for Robust Grasping
\vspace{-10pt}
}

\author{
    \authorblockN{Michael Burgess$^{1}$ and Edward H. Adelson$^{1}$}
        \authorblockA{
     $^{1}$Massachusetts Institute of Technology\\
     {\tt\small \{mburgjr, adelson\}@csail.mit.edu}}
}

\begin{document}

\twocolumn[{%
	\renewcommand\twocolumn[1][]{#1}%
	\maketitle
	\begin{center}
        \vspace{-25pt}
		\includegraphics[width=\textwidth]{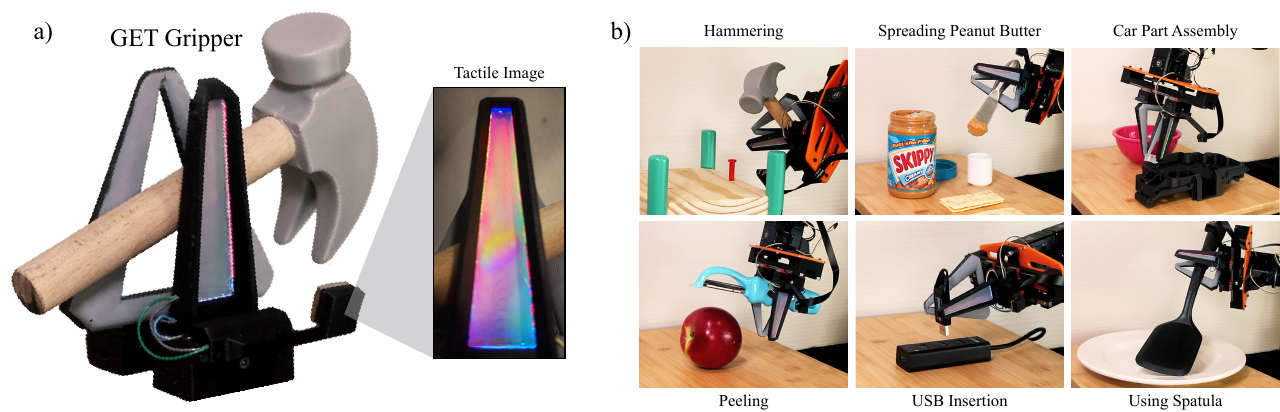}
        \vspace{-18pt}
		\captionof{figure}{\footnotesize{\textbf{Grasp EveryThing (GET) Gripper}. (a) Holding a toy hammer with simultaneous tactile image recorded from HDR camera at the back of the single finger side. (b) The GET gripper is designed to be easily integrated onto parallel grippers commonly found on existing in-the-wild data collection systems with added torque constraint. We demonstrate its capability in completing a variety of household tasks through teleoperation on the ALOHA system~\cite{zhao2023learning}.}}
        \label{fig:teaser}
    \end{center}
}]

\thispagestyle{empty}
\pagestyle{empty}

\begin{abstract}

    We introduce the Grasp EveryThing (GET) gripper, a novel 1-DoF, 3-finger design for securely grasping objects of many shapes and sizes. Mounted on a standard parallel jaw actuator, the design features three narrow, tapered fingers arranged in a two-against-one configuration, where the two fingers converge into a V-shape. The GET gripper is more capable of conforming to object geometries and forming secure grasps than traditional designs with two flat fingers. Inspired by the principle of self-similarity, these V-shaped fingers enable secure grasping across a wide range of object sizes. Further to this end, fingers are parametrically designed for convenient resizing and interchangeability across robotic embodiments with a parallel jaw gripper. Additionally, we incorporate a rigid fingernail for ease in manipulating small objects. Tactile sensing can be integrated into the standalone finger via an externally-mounted camera. A neural network was trained to estimate normal force from tactile images with an average validation error of 1.3~N across a diverse set of geometries. In grasping 15 objects and performing 3 tasks via teleoperation, the GET fingers consistently outperformed standard flat fingers. All finger designs, compatible with multiple robotic embodiments, both incorporating and lacking tactile sensing, are available on \href{https://github.com/GelSight-lab/GraspEveryThing/tree/main}{GitHub}.

\end{abstract}

\begin{figure*}[!t]
    \centering
        \includegraphics[width=\linewidth]{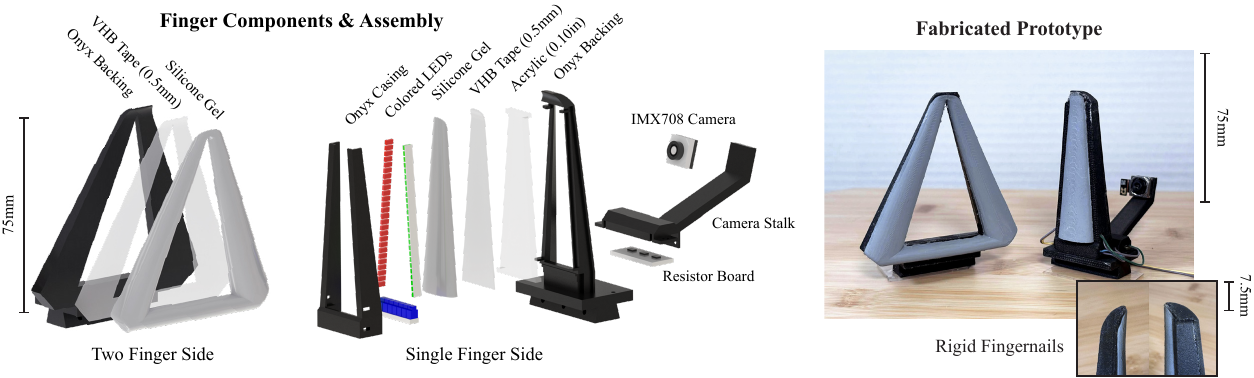}
    \caption{\footnotesize{\textbf{Assembly of GET components}. \textit{Left:} Rendered exploded view of mechanical and electronic components in each finger, arranged in order of assembly. Parts that are made of Onyx are 3D printed on a Markforged printer. \textit{Right:} Fabricated set of GET fingers with rigid fingernails highlighted.}}
    \label{fig:assembly}
    \vspace{-15pt}
\end{figure*}

\section{Introduction}

    Parallel jaw grippers with two flat fingers have dominated robotics in industrial and research settings for decades. These grippers are reliable, easy to control, and simple to model. Recently, their popularity has been reinforced by the rise of learning-from-demonstration style control policies, such as imitation learning and diffusion policy~\cite{chi2023diffusion, fang2019survey, wang2024poco}. These controllers rely on teleoperated demonstrations that are easily collected via one degree-of-freedom (1-DoF) grippers~\cite{chi2024universal, zhao2023learning, aldaco2024aloha}.
    
    Despite their prevalence, traditional flat, two-fingered parallel jaw grippers often create unstable grasps. Often, fingers only make small contact patches on either side of a grasped object. These two antipodal patches cannot strongly resist against torque disturbances~\cite{Prattichizzo2008, carpenter2014comparison}. Therefore, they are not well-equipped to execute dynamic tasks, particularly those requiring tool usage. A more versatile and robust gripper design is needed to match the generality offered by modern learning-based control pipelines.

    By introducing a third finger, we can create a more robust and universally capable gripper that remains as easy to implement and control. The contact patch provided by this additional finger works to form a lever arm to better resist torques about both normal and frictional directions. With V-shaped fingers, the length of this lever varies as needed in correspondence to the shape or size of grasped objects. 
    
    Through maximizing the dynamic capabilities of a 1-DoF gripper, these fingers could be used with existing robotic embodiments and learning-from-demonstration frameworks to expand the range of trainable tasks. Additionally, tactile sensing can be integrated to promote stronger, contact-rich manipulation performance by incorporating additional modalities in data collection beyond traditional vision~\cite{wang2024poco,george2024vitalpretrainingvisuotactilepretraining}.

    We present the GET gripper, a novel design for parallel gripper fingers featuring the following contributions:

    \begin{itemize}
        \item Novel 3-finger design for securely grasping objects and tools of many shapes and sizes.
        \item Rigid fingernails for small objects manipulation.
        \item Integration of camera-based tactile sensing with proximity information in same image and a trained neural network for normal force sensing.
        \item Low cost (less than \$100) and easy to manufacture.
    \end{itemize}


\section{Related Works}

    \subsection{Robotic Gripper Design}
        

        Rigid, parallel jaw grippers are easy to model and control due to their simple geometry and 1-DoF. Recently, they have become especially fundamental in the rise of learning-from-demonstration style control strategies that heavily rely on teleoperation and in-the-wild data collection because they are natural to teleoperate~\cite{chi2024universal, chi2023diffusion}. Despite their convenience, traditional flat-fingered parallel jaw grippers often form only two contact patches on grasped objects. This limits their ability to constrain against torques and achieve secure grasps, making them less effective for dynamic tasks and tool usage~\cite{Prattichizzo2008, manipulation}. Alternative designs have been developed in attempt to overcome this obstacle. Most notable are compliant structure fingers that deform to increase contact area and further geometrically constrain grasped objects~\cite{crooks2016fin, liu2022gelsight, kuppuswamy2020soft}. These grippers are favorable due to their improved performance over rigid fingers, but still fail in dynamic tasks requiring strong torque resistance. Other form factors have been developed to increase torque resistance beyond the traditional 2-fingered parallel gripper~\cite{zhao2023learning,spot}.


        By introducing additional DoF, more dexterous, anthropomorphic, and dynamically-capable grippers can be designed~\cite{piazza2019century, zhao2023gelsight, shaw2023leaphandlowcostefficient, ruotolo2021grasping}. The increased complexity of these grippers empowers more performative action, including strong multi-contact grips and in-hand manipulation behaviors. Yet, hand designs can become bulky, costly, and scale to large numbers of DoF that leave them difficult to control. Crucially, multi-DoF hands are complicated to teleoperate~\cite{fang2025dexo, rohling1993optimized}, which hinders the development of learning-from-demonstration control policies. Naturally, many have sought to minimize DoFs while maximizing dexterity through underactuation~\cite{softrobogrippers}. However, modeling underactuation can be challenging and teleoperation remains nontrivial.

    \begin{figure*}[htbp]
        \centering
            \includegraphics[width=\linewidth]{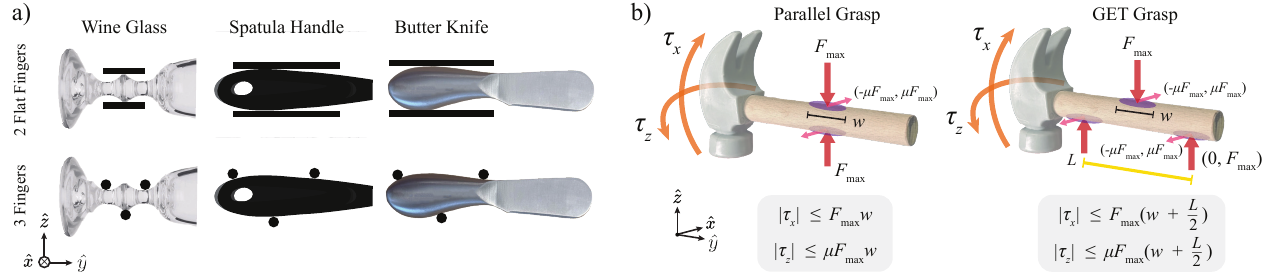}
        \caption{\footnotesize{\textbf{Grasping analysis}. (a) Planar geometric analysis of grasping a set of objects with rigid, 2-finger parallel jaw gripper and our 3-fingered design. (b) Free-body diagram in 3D comparing grasping of a toy hammer with a 2-finger gripper and the GET gripper, where $F_\text{max}$ is the maximum actuation force by the parallel gripper, $\mu$ is the coefficient of friction, $w$ is the width of fingers, $L$ is the distance between fingers, and $\tau_{x,z}$ are disturbance torques applied to the hammer. We take an angular equilibrium about each hammer to calculate the maximum allowable torques before slipping occurs.}}
        \label{fig:grasp_analysis}
        \vspace{-15pt}
    \end{figure*}

    \subsection{Camera-Based Tactile Sensors}

        Camera-based tactile sensors, such as GelSight, record surface deformation through a camera placed behind a soft silicone fingerpad~\cite{yuan2017gelsight,wang2021gelsight}. These sensors can reconstruct the deformed gel surface in 3D due to strict lighting conditions and optical properties of the sensor's design~\cite{yuan2017gelsight,wang2021gelsight,tippur2024rainbowsight,burgess2025learningobjectcomplianceyoungs,willemet2025physicsinformed}. Tactile images from these sensors have been incorporated as a useful modality for learning-based control policies~\cite{Fang2019-dr, george2024vitalpretrainingvisuotactilepretraining, zhao2025polytouch}.
        
        Recently, researchers have developed approaches where a single camera is used to capture both tactile and proximity data, capturing objects and their associated contact geometry in the same image~\cite{zhao2025polytouch, yin2022multimodal}. This eliminates the need for a wrist-mounted camera. However, an externally-mounted camera may be susceptible to lighting changes, introducing noise that could inhibit the ability to reconstruct surface depth as precisely as GelSight sensors traditionally do~\cite{wang2021gelsight}. This is likely acceptable in learning-based implementations that only utilize RGB tactile images as input, not 3D depth~\cite{Fang2019-dr, george2024vitalpretrainingvisuotactilepretraining, zhao2025polytouch}.


\section{Design \& Fabrication}

    \subsection{Mechanical Design}

        \begin{figure}[htbp]
            \vspace{5pt}
            \centering
                \includegraphics[width=\linewidth]{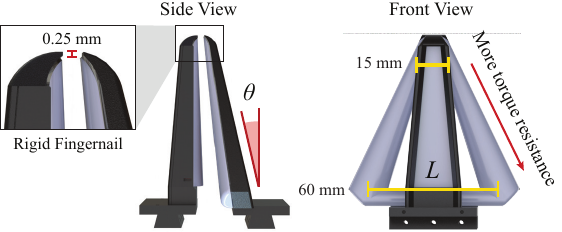}
            \caption{\footnotesize{\textbf{Orthogonal views of GET fingers}. Lever arm of grasp $L$, finger pitch $\theta$, and fingernail dimensions are defined. Smaller objects should be grasped at the tip of the fingers where $L$ is small and larger objects at the base.}}
            \label{fig:design}
            \vspace{-10pt}
        \end{figure}

        \subsubsection{Secure Grasping}

            Traditional parallel jaw grippers with two flat fingers face difficulty in securely grasping convex or irregularly-shaped objects. We analyze the geometry of grasps in 2D over a set of example objects in Fig.~\ref{fig:grasp_analysis}a. Here, we assume fingers to be rigid. The two flat fingers cannot securely grasp these common objects. Forming only two antipodal point contacts, objects can rotate freely out of the plane about the grasping axis $\hat{z}$, even with friction present. The 2-finger gripper also could not strongly resist object rotation about the $\hat{x}$-axis with these point contacts. On the other hand, with the GET gripper's three contact points in a two-against-one configuration, we can better conform to convex or irregular geometries and create grasps that better resist rotation about both the $\hat{x}$ and $\hat{z}$-axes.

            Even when grasping more regularly-shaped objects, traditional parallel jaw grippers with two flat fingers struggle to strongly resist torques. We analyze the forces involved in grasping a hammer with a flat, cylindrical handle in Fig.~\ref{fig:grasp_analysis}b. Here, we assume fingers are not fully rigid and will form patch contacts with the hammer. The 2-finger gripper is able to withstand torque disturbances about the $\hat{x}$ and $\hat{z}$-axes in proportion to the width of the fingers $w$. When a third finger is introduced, the ability to withstand torque additionally depends on the distance between fingers $L$. By designing a parallel gripper with three fingers, we can achieve greater torque compensation about both $\hat{x}$ and $\hat{z}$-axes without the need for overly wide, flat fingers, which may hinder performance in smaller-scale tasks. Resistance to torques about the $\hat{y}$ direction will remain equivalent for both finger configurations and depend on object width.

            With our 3-fingered configuration, it is important to note that the greatest resistance against torque disturbances functions about the $\hat{x}$ axis along the length of the fingers. Resistance against disturbances about the grasping axis $\hat{z}$ will rely on friction, with maximum allowed torque scaling down by the frictional coefficient $\mu$. Objects should be grasped strategically to ensure favorable torque compensation.

            Beyond incorporating a third finger to improve geometric conformity and increase torque resistance, we design fingers with soft silicone gel pads to further enhance these capabilities. When these soft fingerpads come in contact with a grasped object, they will deform to the local geometry of the object. This works to increase the area of contact patches and grants flexibility in grasping irregularly-shaped objects. Additionally, depending on the geometry of the object, an indentation formed at the contact patch may provide slip resistance along the face of the finger, with the elastic silicone acting as a spring to resist object movement.

        \begin{figure*}[!t]
            \centering
                \includegraphics[width=\linewidth]{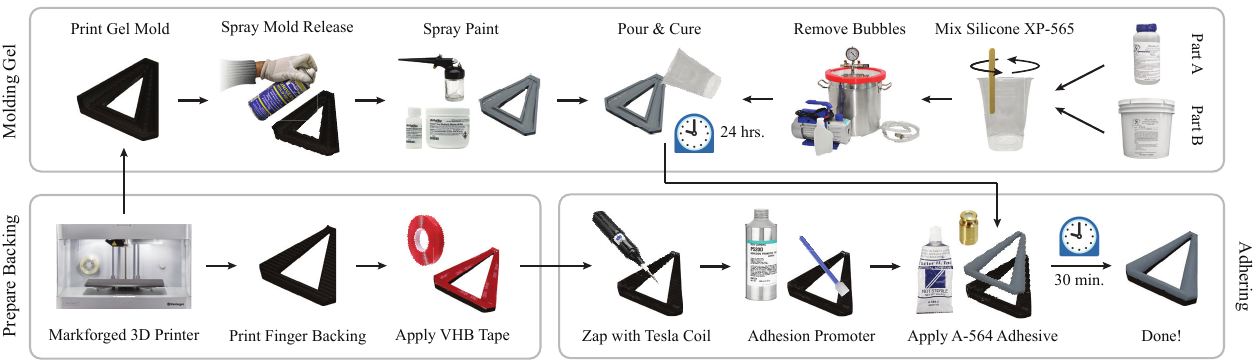}
            \caption{\footnotesize{\textbf{Fabrication process of GET fingers}. Required steps to fabricate designed fingers, including molding of silicone gels, preparing plastic backings, and adhering soft silicone fingerpads to backings. In total, process takes multiple days to complete with bulk of the time dedicated to silicone curing.}}
            \label{fig:fabrication_process}
            \vspace{-10pt}
        \end{figure*}

        \subsubsection{Self-Similarity \& Scalability}\label{sec:scaling}

            Our gripper design is inspired by the principle of self-similarity~\cite{hutchinson1981fractals}. In the application of manipulation, we aim to design a gripper that can interact with objects of many sizes and be seamlessly interchanged across different robotic embodiments.

            On the 2-finger side of our gripper, the fingers come together to form a V-shape. This creates a varying distance $L$ between fingers along their length to enable secure grasping of many object sizes, as shown in Fig.~\ref{fig:design}. Larger $L$ at the base provides greater compensation against torque disturbances, while smaller $L$ at the tip enables grasping of small objects. With this design, it is expected that larger objects are grasped at the base of the fingers and smaller objects at the tip.

            The thickness of each finger varies along its length. The finger backing is thickest and most stiff at its base, becoming thinner and more compliant towards the tip. Similarly, gel thickness varies from base to tip, enabling more conformity at the base for larger objects. This design is inspired by recent compliant-structure fingers~\cite{crooks2016fin,liu2022gelsight}. 
            
            To prevent fingers from bending outward and diverging, which could lead to grasp instability, we pitch them inward by angle $\theta$. Importantly, fingers are designed to interdigitate. They can bend flat to a $0^\circ$ pitch angle through contact without interfering with each other, allowing even infinitesimally thin objects to be grasped along the length of the finger.

\begin{figure}[htbp]
                \centering
                    \includegraphics[width=0.8\linewidth]{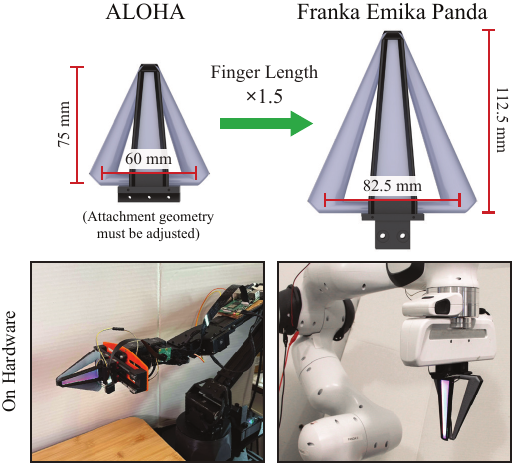}
                \caption{\footnotesize{\textbf{Scalable finger design}. Designs are parametrized by finger length for use across robotic embodiments. To demonstrate this flexibility, fingers are fabricated for both the ALOHA system~\cite{zhao2023learning} and Franka Emika Panda arm.}}
                \label{fig:scaling_fig}
                \vspace{-10pt}
            \end{figure}

            Fingers are parametrically designed such that they can be resized for compatability with any robotic arm. This is done by adjusting finger length dimension and attachment geometry. Throughout this paper, we refer specifically to fingers designed for the ALOHA system~\cite{zhao2023learning}. But, to demonstrate scalability, additional desgins were fabricated for use with a Franka Emika Panda arm, as shown in Fig.~\ref{fig:scaling_fig}, and the UMI gripper, as available on \href{https://github.com/GelSight-lab/GraspEveryThing/tree/main}{GitHub}~\cite{chi2024universal}.

        \subsubsection{Rigid Fingernail}

            A rigid fingernail is incorporated at the tip of each finger to enhance ability in grasping small objects. The fingernail assists in prying objects from their surroundings (e.g., a flat surface or a pile of clutter) and lifting them into the gripper's fingers.  Our fingernail design seamlessly integrates with the rest of the finger, smoothly transitioning into the soft fingerpads. This ensures that the fingernails do not impede the gripper's ability to grasp larger objects. This design is inspired by previous robotic fingers that have integrated fingernails~\cite{jain2020soft, kumagai2023improvement, li2024underactuatedroboticgrippermultiple}. The use of these fingernails for grasping a coin and paper clip is demonstrated in Fig.~\ref{fig:fingernail_grasping}. 

    \begin{figure*}[!t]
        \centering
            \includegraphics[width=\linewidth]{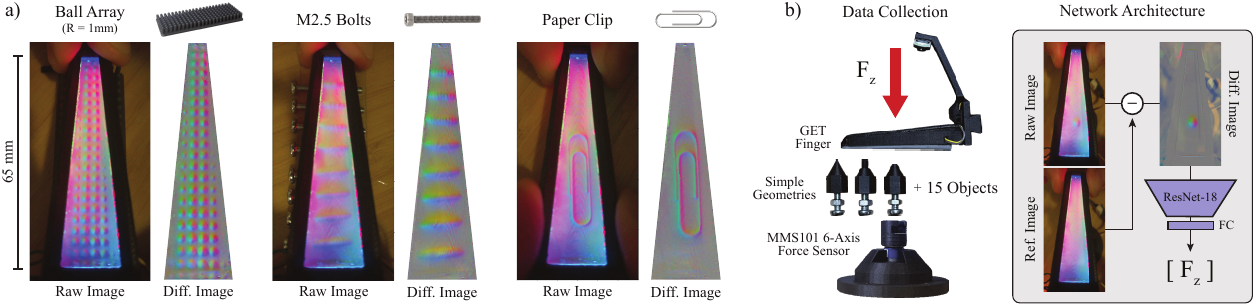}
        \caption{\footnotesize{\textbf{Integrated tactile sensing in GET finger}. (a) Tactile images recorded from contact with a set of objects. Difference images are calculated with respect to an uncontacted reference image. (b) \textit{Left:} Data collection process using force probe with interchangeable indentation geometries, including 3 simple geometries and 15 common objects. \textit{Right:} Force sensing architecture where difference images are sent through ResNet-18 to estimate normal force $F_z$. }}
        \vspace{-15pt}
        \label{fig:tactile}
    \end{figure*}

    \subsection{Optical Design}

        Camera-based tactile sensing is integrated into the single finger of the gripper. Soldered arrays of RGB LED are placed along the edges of the finger to illuminate the sensor pad. Our finger is significantly longer than it is wide. In compensation, lensed blue LEDs are used at the base of the finger to direct light further along its length. Otherwise, red and green SMD LEDs are placed on either side of the finger and equipped with diffusion filters to promote uniform illumination. 
        
        As discussed in Section~\ref{sec:scaling}, our design is intended to be parametrically scalable. In accordance with this, each LED board could be replaced by flexible COB LED strips, which can be customized to different lengths, eliminating the need to design a new PCB board for each new finger size.

        Gel pads are 4.5~mm deep at the base of the finger. This creates a steep incident angle between the LEDs and the gel surface. Considering this, the gel is coated with gray Lambertian paint to evenly scatter light and improve uniformity.
    
        A camera is mounted behind the finger on a stalk, angled toward the center of the backing. The backing of the single finger remains open and optically clear, allowing the camera to view the surface of the gel pad. Unlike traditional GelSight sensors~\cite{yuan2017gelsight, wang2021gelsight}, we do not enclose our sensor's camera within an opaque housing. This enables the camera to simultaneously capture environmental proximity information and tactile data in the same image stream. To ensure that background details are not washed out by the bright illumination required for our sensor, we use a high dynamic range (HDR) camera.

    \subsection{Fabrication Process}
    
        The following steps are taken to fabricate fingers. In our experience, this process creates fingers that are capable of enduring at least 1000 usage cycles without any noticeable signs of wear.

        \subsubsection{Prepare Backings}

            Backings for each finger are printed on a Markforged 3D printer with Onyx, a reinforced nylon composite. Acrylic pieces are laser cut from a 0.10~in thick sheet and super-glued into place on the single-finger backing. Afterward, a layer of 0.5~mm clear 3M™ VHB™ tape is placed and firmly pressed onto the surface of each backing and cut to size. Fabrication processes are shown in Fig.~\ref{fig:fabrication_process}

        \subsubsection{Molding Gels}

            Molds for each finger's gel are printed on a Markforged 3D printer. Since gel thickness varies along the length of each finger, molds retain fine, ring-like patterns that partially resemble a human fingerprint, an artifact of layer-height from the printing process. These textures may work to increase friction on the gel surface. Patterns may be seen on the fabricated prototype in Fig.~\ref{fig:assembly}.

            Print-On® gray ink base is combined with silicone ink catalyst in a 10:1 mass ratio to form a Lambertian paint mixture. Each mold is sprayed with Smooth-On Universal™ Mold Release before being airbrushed with a thin, opaque layer of paint. Optically clear Silicones Inc. XP-565 is mixed in an 18:1 ratio, degassed, and poured into the mold. Cured gels are carefully removed from the mold after 24-48~hrs.

        \subsubsection{Adhering}

            With gels and backings prepared, the surface of VHB tape is zapped with a tesla coil. Then, a thin layer of DOWSIL™ P5200 Clear Adhesion Promoter is spread across the surface. After 1-2~min, another layer is applied. Immediately after that, A-564 Medical Adhesive is spread over the tape. Gels are then placed on top with small weights to ensure a secure bond. Leave to cure for 20-30~min.

        \subsubsection{Lighting}

            Blue 1.8~mm lensed LEDs are soldered onto a 0.6~mm thick PCB board sized by finger width. 0.75~mm thick red and green SMD LEDs (1206 package) are soldered onto 0.6~mm thick PCB boards sized by finger length. All LED arrays are connected to a small PCB board with resistors of 45, 400, and 600~$\Omega$ for red, green, and blue respectively. This board is connected to a 3.3~V power source. Resistor values are calibrated empirically to maximize uniformity of illumination for specific finger geometry and LEDs.
            
            All LED boards are super-glued onto a 3D printed front casing. This casing attaches to the backing via press-fit and can be further secured with super-glue. Before the casing is put into place, gray and diffusion filters are applied along the sides of the acrylic. The gray filter used is ``VViViD Smoke Black Adhesive Headlight Wrap'', while the diffusion filter used is 3M™ Diffuser 3635-70~\cite{wang2021gelsight}.

        \subsubsection{Camera}

            The camera stalk is printed on a Markforged 3D printer and attached to the base of the single finger with M2 screws. This stalk also serves to cover the PCB resistor board. An IMX708 camera is attached to the top of the stalk using 3M™ VHB™ tape. The camera and LEDs are connected to a Raspberry Pi 4B.
    
    \begin{figure*}[!t]
        \centering
            \includegraphics[width=0.7\linewidth]{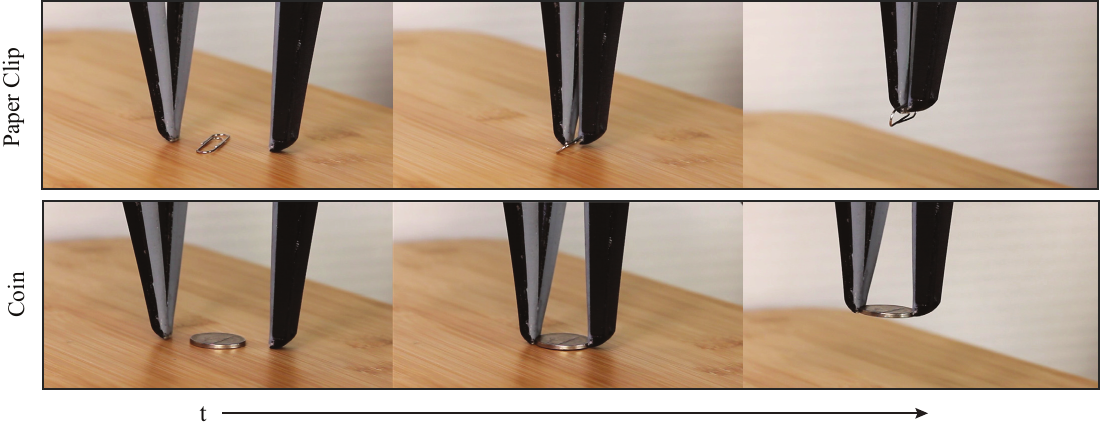}
        \caption{\footnotesize{\textbf{Grasping small objects with rigid fingernail}. The process of grasping small objects, such as a coin and paper clip, off of a flat surface with GET fingers. The nail is used to pry underneath the object and lift it off the table. Grasping is executed via teleoperation on the ALOHA system~\cite{zhao2023learning}.}}
        \vspace{-15pt}
        \label{fig:fingernail_grasping}
    \end{figure*}

    All CAD files and code for streaming tactile images from the sensor are available on \href{https://github.com/GelSight-lab/GraspEveryThing/tree/main}{GitHub}. We upload CAD files for versions of the fingers compatible with the ALOHA system~\cite{zhao2023learning}, Franka Emika Panda arm, and UMI gripper~\cite{chi2024universal}. Additionally, CAD files are uploaded for versions of the fingers without integrated tactile sensing for a more lightweight design in applications where sensing is not desired. 
    

\section{Experiments \& Results}

    \subsection{Tactile Sensing}

        Tactile images were captured using an externally-mounted IMX708 camera. Due to the camera's high dynamic range (HDR), proximity information can be captured in the same image as tactile information. Example tactile images are shown in Fig.~\ref{fig:tactile}a. Fine surface features, such as the threads of an M2.5 bolt, are captured in these images. However, it is difficult to uniformly focus the camera along the entire length of the finger. In the provided images, top and bottom regions are more in-focus than the center, resulting in sharper images with higher clarity in those areas. Opposing red and green lights provide balanced illumination, while the lensed blue LEDs at the base effectively distributes light to the fingertip. Due to the curved and irregular profile of the gel pad, achieving uniform color mixing remains challenging, particularly at the edges of the sensor.

        Tactile images were collected alongside corresponding measurements from a force probe over varying indentation geometries. Indentation geometries include 3 simple shapes and 15 common objects, such as a coin, USB stick, Lego brick, and more. The force probe used for this experiment was designed with an embedded MMS101 6-axis force torque sensor. CAD files for this probe are available on \href{https://github.com/GelSight-lab/GraspEveryThing/tree/main}{GitHub}. To collect data, we pressed our single finger tactile sensor into the probe multiple times for each geometry while streaming tactile images and recording force measurements for each frame, as shown in Fig.~\ref{fig:tactile}b. Pressing forces ranged from 0 to 30~N over the collected dataset. In total, 17.5k images were collected. For reference, while streaming at 30~Hz, all of this data can be collected in under 10~min.

        Collected data was used to train a neural network to estimate normal force $F_z$ from tactile difference images. This architecture was based on \cite{Azulay_2024}, with ResNet-18 as its backbone. We trained over data collected from simple indentation geometries and 15 common objects. Models were trained for 50 epochs with a batch size of 32, learning rate of 0.001, and the same random seed with 20\% of total data used for validation. Trained models were evaluated over a set of 3 unseen objects.

        The resulting errors from our model during training and over unseen objects are provided in Table~\ref{table:force_sensing}. We observed accurate force estimation for each indentation geometry and a strong ability to generalize across object shapes. Our sensor was not fabricated with markers. Force sensing with GelSight-like sensors can be improved by painting black markers on the gel pad surface to track surface displacement~\cite{yuan2017gelsight}.
        
        \begin{table}[h]
            \small
            \begin{minipage}{\linewidth}
            \caption[font=small,labelfont=bf]{Force Sensing Results}
            \vspace{-5pt}
            \centering
                    \begin{tabular}{cccc}
                    \toprule
                    Shape(s) & Images & Train RMSE & Val. RMSE \\ 
                    \midrule
                    Simple Geom. & 4.5k & 0.46~N & 1.12~N \\ 
                    Simple Geom. + 15 Obj. & 17.5k & 0.52~N & 1.34~N \\ 
                    \midrule
                    M5 Nut \textit{(Unseen)} & 800 & — & 1.40~N \\ 
                    Golf Ball \textit{(Unseen)} & 800 & — & 2.65~N \\ 
                    Screwdriver \textit{(Unseen)} & 800 & — & 2.83~N \\ 
                    \bottomrule
                    \label{table:force_sensing}
                    \end{tabular}
            \end{minipage}
            \vspace{-15pt}
        \end{table}

    \subsection{Performance in Grasping Objects}\label{sec:obj_results}

        We evaluated our fingers' ability to grasp a set of 15 objects in comparison to standard parallel jaw fingers. To assess hardware independent of controllers, all robotic actions were teleoperated via the ALOHA system~\cite{zhao2023learning}. Objects were chosen among common household objects that encompass a wide range of shapes and sizes. Objects were selected with respect to the ALOHA system's payload capacity of 0.75~kg~\cite{zhao2023learning}.

        For these experiments, we utilized the ALOHA system~\cite{zhao2023learning} retrofitted with an enhanced gripper design from the ALOHA 2~\cite{aldaco2024aloha}. We compared three different finger configurations: our proposed design, ALOHA ViperX fingers, and a custom-designed set emulating traditional parallel jaw fingers. The traditional fingers are flat and rectangular with a rigid backing and a soft curved gel pad that mimics GET fingers. The ALOHA ViperX fingers are fully rigid with a triangular design that can provide up to four points of contact. We applied high-friction tape across the entire surface of ALOHA ViperX fingers to maximize their capabilities.

        \begin{figure*}[!t]
            \centering
                \includegraphics[width=\linewidth]{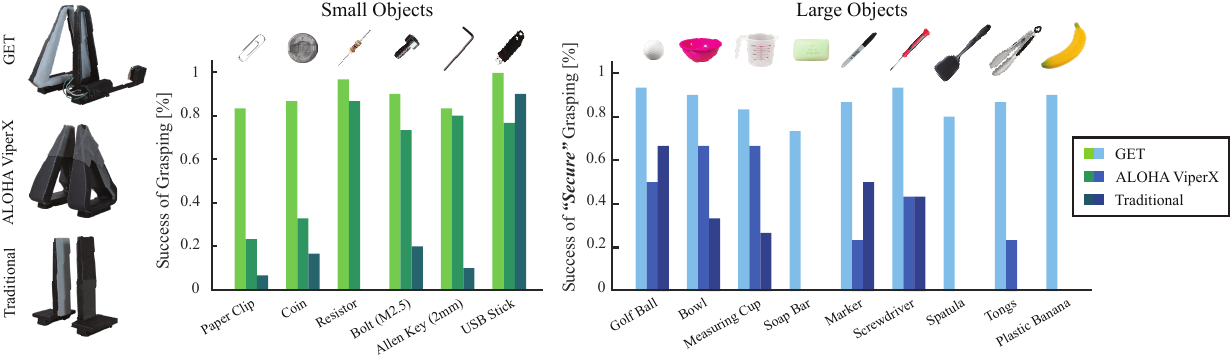}
                \vspace{-22pt}
            \caption{\footnotesize{\textbf{Performance in grasping objects}. 15 objects of varying size and shape were grasped using GET, ALOHA ViperX~\cite{zhao2023learning}, and traditional parallel jaw fingers. Success in grasping small objects is defined as the ability to grab them off a table. Success in grasping large objects is defined as the ability to \textit{securely} grasp objects without slipping when subjected to disturbance of \textless3~N in any direction. Results are reported over 30 trials per object.}}
            \label{fig:obj_results}
        \end{figure*}

        \begin{figure*}[!t]
            \centering
                \includegraphics[width=\linewidth]{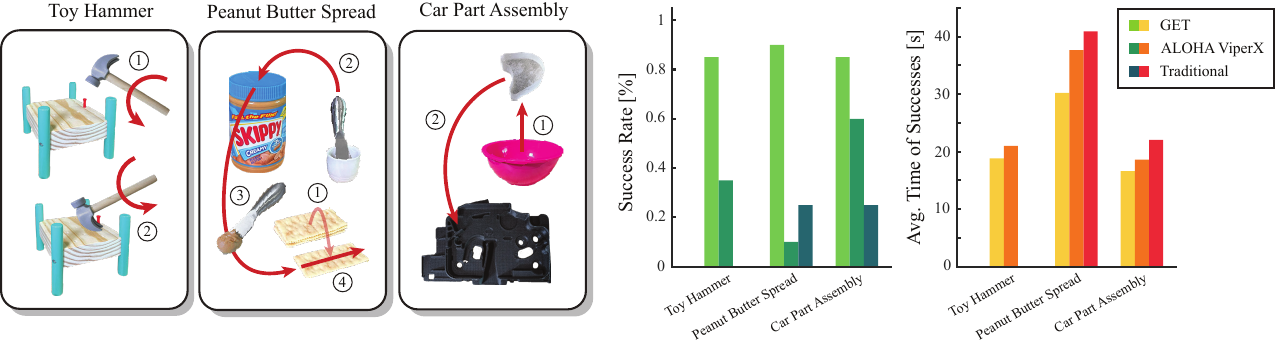}
                \vspace{-20pt}
            \caption{\footnotesize{\textbf{Performance in manipulation tasks}. 3 manipulations tasks, as defined on the left, were completed using GET, ALOHA ViperX~\cite{zhao2023learning}, and traditional parallel jaw fingers. Completion time is defined as the average over successful trials for each gripper. Results are reported over 20 attempts per task.}}
            \label{fig:task_results}
            \vspace{-10pt}
        \end{figure*}

        Teleoperation was performed by a set of 3 operators of varying skill levels. Results are reported from each of these teleoperators in aggregation. Most objects started flat on the center of a table in a random orientation, with the exception of larger tool objects that began vertical inside of a cup. Operators attempted to grasp each object with each gripper for 10 trials. Between every trial, the scene was fully reset. Objects were grasped with maximum force allowed by the ALOHA system, which is measured up to approximately 15~N.

        Selected objects are categorized as either ``small'' or ``large'', with grasp success measured differently for each. For small objects, a grasp was considered successful if the teleoperator could grab the object off of the table and hold it in the air for 5~s without slipping. For large objects, it was assumed that all could be grasped off the table by each gripper, thus success was instead measured by grasp security. A grasp was considered ``secure'' if the object did not slip or dislocate when subjected to disturbance forces of up to 3~N applied in any direction. To measure this, grasped objects were probed from multiple directions using the force probe, shown in Fig.~\ref{fig:tactile}b, while force recordings were monitored and slip was determined visually.

        Grasping success for all fingers across large and small objects is plotted in Fig.~\ref{fig:obj_results}. The GET fingers grasped small objects more successfully than baseline fingers. This was likely due to the use of our fingernail design to pick small objects off of the table, as demonstrated in Fig.~\ref{fig:fingernail_grasping}.

        Furthermore, the GET fingers were more successful in securely grasping large objects. As shown in Fig.~\ref{fig:grasp_analysis}a, traditional parallel jaw fingers create only two antipodal contact patches and must rely on friction to resist disturbances about the grasping axis. This friction is often minimal given small contact patches, resulting in a weak grip where objects are only loosely held in their orientation. The ALOHA ViperX fingers faced similar challenges. Although these fingers can theoretically provide up to four contact points, their rigid profile could not conform to convex or asymmetric geometries, often resulting in contact only being formed at two points. Consequently, they behaved similarly to traditional flat fingers.
        
        In contrast, the GET fingers more securely grasped large objects. With their ability to create three contact patches and adapt to object geometries, they could better resist disturbances about the normal and frictional directions for all objects. Moreover, when an object is grasped, it indents into the soft gel pads, creating a contact patch that provides elastic resistance in tangential directions. This reduces reliance on friction and discourages slipping due to the gel's spring-like behavior. As a result of these contributions, the GET fingers significantly outperformed baseline fingers in securing large objects.

    \subsection{Performance in Manipulation Tasks}

        We evaluated our gripper's ability to complete a set of 3 tasks in comparison to the ALOHA ViperX gripper and a custom-designed traditional fingers. For this experiment, the setup was nearly identical to that of Section~\ref{sec:obj_results}. Tasks were teleoperated via the ALOHA system using a single arm~\cite{zhao2023learning}.

        Tasks were designed to assess grippers' ability to securely hold objects through torque disturbances and manipulate small objects. All tasks were defined as follows:
        \begin{itemize}
            \item \textit{Toy Hammer:} A hammer was placed vertically on the table next to a fixed-position board with a nail loosely in a hole. The hammer must be grabbed and used to hit the nail into the board. After, the nail must be pried back out of the toy board using the other side of the hammer. During this process, the hammer should not dislocate or slip from the grasp. 
            \item \textit{Peanut Butter Spread:} A peanut butter jar with a loose lid was placed next to a stack of crackers. The lid must be removed from the jar. Then, a single cracker must be picked and moved to the edge of the table. A butter knife must be grabbed from a vertical orientation, used to scoop peanut butter from the jar, and spread it across the length of the cracker. During this process, no objects should be dropped and the knife should not dislocate or slip from the grasp. 
            \item \textit{Car Part Assembly:} A pile of small rubber pieces inside a bowl was randomized between each trial. A piece must be picked from the pile and transported into its place in the fixed-position plastic piece without being dropped.
        \end{itemize}

        Task performance for each set of fingers is plotted in Fig.~\ref{fig:task_results}. Our fingers outperformed baseline fingers, specifically in the Toy Hammer and Peanut Butter Spread tasks. Traditional fingers struggled to resist torque moments in these tasks given only two small contact patches. Furthermore, they could not reliably pick small objects in the Car Part Assembly task. ALOHA ViperX fingers performed better in the Car Part Assembly task, but struggled in the Peanut Butter Spread task. These rigid fingers could not conform to the butter knife's convex geometry to securely grasp it, as displayed in Fig.~\ref{fig:grasp_analysis}a.

        The GET fingers more successfully completed all tasks than baseline fingers. In the Car Part Assembly task, their fingernails helped pick small pieces from clutter and place them firmly into the desired location. In the Peanut Butter Spread task, the GET design could securely grasp the irregularly-shaped butter knife through contact with the cracker. Fingernails were also helpful in this task for removing the peanut butter lid and picking crackers from the stack. In the Toy Hammer task, GET fingers outperformed the ALOHA ViperX fingers, even with the hammer's flat, cylindrical geometry. GET fingers had greater capacity to resist high torque required to pry the nail from the board, as shown in Fig.~\ref{fig:grasp_analysis}b. In all tasks, we observe decreased completion time with the GET fingers, suggesting greater teleoperator confidence and ease of use.

    \subsection{Limitations \& Future Work}
    
        Despite its capabilities, our gripper, like any other 1-DoF gripper, cannot perform in-hand manipulation or reorientation. To partially overcome this limitation, a bimanual manipulation interface can be utilized to complete further dexterous tasks. Additionally, due to the gripper's asymmetry, a robotic arm with at least 7-DoFs is required to reach any arbitrary pose, unlike standard symmetric grippers that only need 6-DoFs.

        Our tactile sensor does not currently track contact geometry in 3D. Traditionally, GelSight sensors use color mapping to compute gradients and reconstruct depth across the contact surface~\cite{wang2021gelsight}. Since our sensor is exposed to external lighting conditions, we cannot rigorously develop this same mapping. For the time being, we forgo depth estimation. If deemed necessary, future work could explore training a neural network to reconstruct depth from tactile images. Additionally, a neural network could be trained for multi-axis force sensing by incorporating markers on the gel pad surface.


\section{Conclusion}

    We present the GET gripper, a novel 3-fingered design for parallel jaw actuators to more securely grasp objects of many shapes and sizes. This design has been shown to outperform standard 2-fingered designs, particularly in grasping small objects and securely holding larger ones, such as tools for dynamic tasks. Additionally, high-resolution tactile imaging is integrated into fingers with force sensing capabilities. The GET gripper could be deployed on teleoperation interfaces to collect rich data for training policies over a wide range of tasks. Designs for use with ALOHA, Franka Emika Panda, and UMI systems are available on \href{https://github.com/GelSight-lab/GraspEveryThing/tree/main}{GitHub}, including variants without tactile sensing or silicone gels for rapid manufacturing.




\section*{ACKNOWLEDGMENTS}
    
    Toyota Research Institute (TRI) provided funds to support this work. Thank you to Megha Tippur for help with PCB board design and Yuxiang Ma for setting up ALOHA system.


\bibliographystyle{IEEEtran}
\bibliography{refs.bib}

\end{document}